\documentclass[11pt]{article}

\usepackage[preprint]{acl}

\usepackage{times}
\usepackage{latexsym}

\usepackage[T1]{fontenc}

\usepackage[utf8]{inputenc}

\usepackage{microtype}

\usepackage{inconsolata}

\usepackage{graphicx}
\usepackage{amsmath}
\usepackage{amssymb}
\usepackage{tikz}
\usetikzlibrary{positioning,decorations.pathreplacing,calc}
\title{Generative-Evaluative Agreement: A Necessary Validity Criterion for LLM-Enabled Adaptive Assessment}

\author{Grandee Lee \and Yue Wang \and Che Yee Lye \and Luke Peh
 \\
  Singapore University of Social Sciences, \\463 Clementi Rd, 599494, Singapore \\
  \texttt{\{grandeelee,wangyue,cylye,lukepehlc\}@suss.edu.sg} \\}

\begin{document}
\maketitle
\begin{abstract}
When the same LLM generates assessment items, simulates student responses, and scores them, the validation loop is self-referential. We introduce \textbf{Generative-Evaluative Agreement (GEA)}, a validity criterion measuring whether an LLM's scoring function recovers the skill levels its generative function was instructed to produce. In the first direct measurement of GEA on a two-stage adaptive assessment, the model recovers roughly half the intended variance ($r = 0.698$) with systematic positive bias. GEA is strong ($r > 0.7$) for syntactically verifiable skills but near zero for design-level skills, and low-skill overestimation inflates scores near the routing threshold. We argue that granular, skill-decomposed rubrics are the principal proposed mechanism for strengthening GEA and outline complementary mitigations.
\end{abstract}
\section{Introduction}

Computerized adaptive testing (CAT) traditionally relies on item banks pre-calibrated via Item Response Theory (IRT), where every item has known difficulty and discrimination parameters estimated from hundreds of real responses \citep{vanderLinden2010}. LLM-enabled adaptive assessment disrupts this paradigm: items are generated dynamically, each student potentially receives a unique test, and classical calibration (requiring 50--200 respondents per item; \citealt{Lord1980}) becomes infeasible. Revisions to rubrics, prompts, or course materials become psychometrically consequential when they change what is being measured, how performance is elicited, how responses are scored, or how scores are interpreted. In these cases, prior calibration and validity may no longer be transportable, and at least part of the item bank may need to be re-authored, relinked, re-calibrated, or revalidated \citep{Han2011}. In a typical school setting, learning outcomes are updated when curricula evolve, rubrics are refined each semester as instructors identify ambiguities, and course restructuring changes the skill prerequisites for each assignment. LLM-based systems absorb these changes through prompt and rubric updates alone, but the validity of the resulting assessment must be re-established each time.

This creates a \textbf{bootstrapping problem}: the system cannot be validated without real student data, but cannot be deployed at scale without prior validation. Human review of each generated item is infeasible when the item space is effectively infinite, and the classical validity pipeline (pre-calibrate, validate against human raters, then deploy) does not apply. Simulation-based validation offers a pragmatic alternative. \citet{Liu2025} demonstrated that ensembles of LLM-simulated respondents can approximate human item calibration with correlations exceeding 0.89. \citet{Zheng2025} used Monte Carlo simulation to identify optimal CAT configurations before empirical evaluation. \citet{MarquezCarpintero2024} reviewed LLM-simulated student profiles for pre-deployment testing of pedagogical systems. However, when the same LLM generates items, simulates student responses, and scores them, the validation loop is self-referential. If the model's representation of skill levels is inconsistent across its generative and evaluative functions, the system validates itself against a distorted mirror.

This paper introduces \textbf{Generative-Evaluative Agreement (GEA)} as the formal criterion for this internal consistency: when an LLM generates a response at an intended skill level, does scoring recover that level? Valid routing decisions require scores that faithfully reflect the intended construct, but ``intended difficulty'' exists only in the model's internal representation, accessed through two different computational paths (generation and evaluation) that traverse different prompt-conditioned regions of the same model. Empirical verification of their alignment is therefore a necessary (though not sufficient) validity condition for any LLM-based adaptive assessment that uses simulation for calibration.

\begin{figure}[t]
	\centering
	\includegraphics[width=\linewidth]{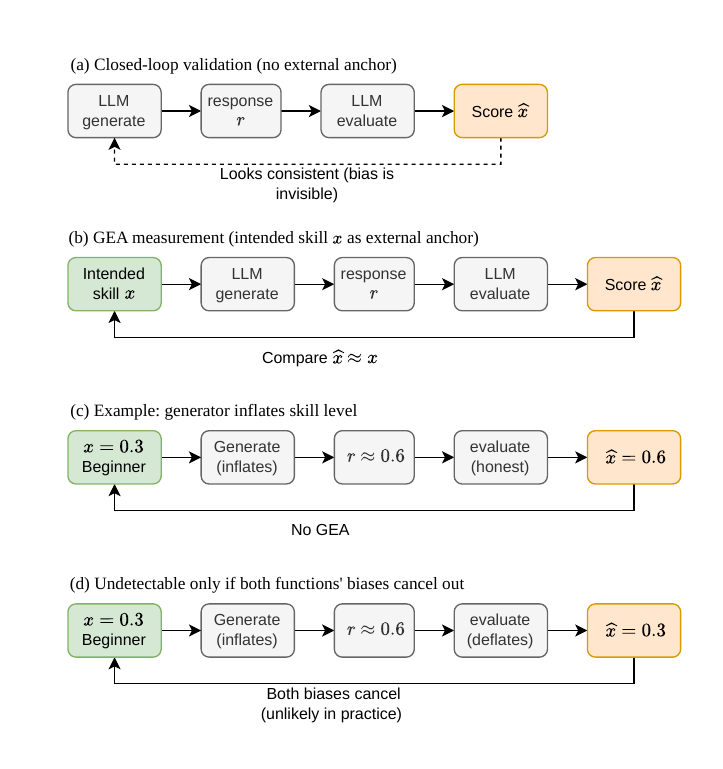}
	\caption{GEA measurement versus closed-loop self-validation. (a)~In a pure closed loop, there is no external anchor and bias is invisible. (b)~GEA introduces the intended skill level $x$ as an external reference point. (c)~When the generator inflates skill, GEA detects the discrepancy. (d)~Bias is undetectable only if both functions share the exact same misconception, an unlikely scenario given that generation and evaluation traverse different prompt-conditioned paths.}
	\label{fig:gea-vs-closedloop}
\end{figure}

\subsection{Definition}
\label{sec:definition}

\textbf{Generative-Evaluative Agreement (GEA)} is the degree to which an LLM's generative representation of skill levels is consistent with its evaluative representation. Formally, if the model generates a response $\mathbf{r}$ conditioned on skill level $x$, then scoring $\mathbf{r}$ should recover $x$ within acceptable error bounds:
\begin{equation}
\label{eq:gea}
\mathbb{E}[\mathrm{score}(\mathbf{r}) \mid \mathbf{r} \sim \mathrm{generate}(x)] \approx x
\end{equation}
Here $x, \mathrm{score}(\mathbf{r}) \in [0,1]$ are continuous per-skill scores; ordinal proficiency bands (Appendix~\ref{sec:appendix-scale}) are derived post hoc for reporting.

We operationalise ``$\approx$'' through two primary metrics: Pearson $r$ for rank-order fidelity and signed bias for systematic directionality. We propose two actionable benchmarks: $r > 0.7$ (strong GEA) to support fine-grained proficiency reporting, and $r > 0.4$ (moderate GEA) to support binary routing decisions. Skills below $r = 0.4$ should not be used for adaptive routing without human validation.

\begin{figure*}[htbp]
	\centering
	\includegraphics[width=.7\linewidth]{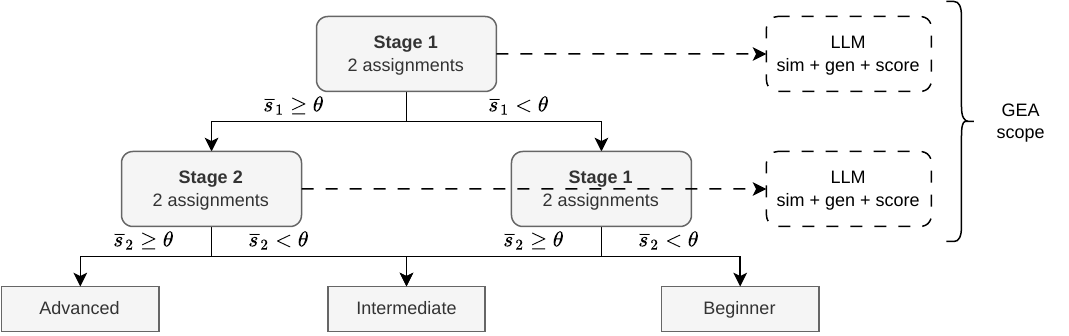}
	\caption{Adaptive assessment flow. The LLM generates assignments and scores responses at each stage. Routing depends on cumulative score $\bar{s}$ crossing threshold $\theta$. GEA measures consistency between the LLM's generative and evaluative functions.}
	\label{fig:assessment-flow}
\end{figure*}

Crucially, GEA measurement is \emph{not} equivalent to closed-loop self-validation. Figure~\ref{fig:gea-vs-closedloop} illustrates the distinction. In a pure closed-loop system (panel~a), the model generates and scores with no external reference, so any systematic bias is invisible. In GEA measurement (panel~b), the intended skill level $x$ serves as an external anchor. If the generator inflates skill (panel~c), the discrepancy $\hat{x} \neq x$ reveals the generation bias. GEA can only fail to detect bias when both functions share the \emph{exact same} misconception of $x$ (panel~d), which is unlikely given the empirical evidence of divergence reviewed in Section~\ref{sec:empirical-divergence}. Figure~\ref{fig:assessment-flow} shows the concrete assessment architecture in which GEA is measured.

\section{Background}
\label{sec:background}

\subsection{The Closed-Loop Problem}
\label{sec:closed-loop}

In traditional CAT, item parameters and scoring functions are independently validated against real human response data. In LLM-based adaptive systems, the model performs both roles with no external anchor. An instructive (though imperfect) analogy comes from the generative/discriminative distinction: \citet{Ng2002} showed that models learning $P(X \mid Y)$ and $P(Y \mid X)$ can disagree at finite capacity. In an LLM, both generation and evaluation share the same weights, but are conditioned on different prompts that traverse different computational paths. Generation is dominated by fluency priors; evaluation by criterion matching. Shared architecture makes alignment \emph{plausible} but does not \emph{guarantee} it \citep{Oh2024,West2023}.

\subsection{Empirical Evidence of Divergence}
\label{sec:empirical-divergence}

LLMs struggle to simulate lower-proficiency cognitive states \citep{Yuan2026}: expert knowledge leaks through despite skill-level prompting. \citet{Srivatsa2025} tested 11 LLMs against real NAEP data and found no model-prompt pair faithfully reproduced real student distributions; \citet{Wu2025} confirmed this for Python programming. LLMs also systematically rate their own outputs higher than equivalent text from other sources (\textbf{self-preference bias}; \citealt{Panickssery2024}), with the mechanism identified as perplexity-based familiarity \citep{Wataoka2024}. Even proprietary models show low intra-rater consistency at temperature $> 0$ \citep{Wang2024consistency}. In simulation-based calibration, these mechanisms compound: the result may appear internally consistent but is not externally valid.

\subsection{Implications for Calibration}
\label{sec:calibration-implications}

When simulation is the only feasible calibration method, GEA becomes the gatekeeper for trustworthiness. If GEA is low, score distributions reflect the model's self-consistency rather than real student performance. \citet{Rahim2025} argue that Generalizability Theory and Many-Facet Rasch Measurement are needed to decompose multiple simultaneous error sources rather than collapsing them into a single coefficient. Even in real deployment, the generative side affects question generation: if items are at the wrong difficulty, routing decisions are based on mis-targeted items regardless of grading accuracy.

\subsection{Related Work}
\label{sec:related-work}

GEA connects to several research threads. The automated essay scoring (AES) literature has studied inter-rater reliability for decades \citep{Ramesh2022}; GEA differs in that rater and author are the same model. From a psychometric perspective, GEA instantiates the \emph{substantive} component of \citeauthor{Messick1989}'s (\citeyear{Messick1989}) construct validity framework, and aligns with the \emph{Standards for Educational and Psychological Testing} \citep{AERA2014} requirement for evidence that scores support intended interpretations. The LLM-as-judge paradigm \citep{Zheng2024} has documented self-preference and position bias; GEA extends this from evaluation-only settings to the generate-then-evaluate pipeline where generation bias compounds with evaluation bias.

\section{Empirical Measurement of GEA}
\label{sec:empirical}

This section presents the empirical measurement of GEA in the sense defined in Section~\ref{sec:definition} for the concrete case of Python object-oriented programming (OOP) coding tasks. The same Claude model performs both code generation and rubric-based evaluation against the 24-skill taxonomy (class definition, inheritance, exception handling, etc.; full list in Appendix~\ref{sec:appendix-skills}).

\subsection{Simulation Design}
\label{sec:sim-design}

\paragraph{Student profiles.}
We generated 150 synthetic student profiles, each comprising a
24-dimensional skill vector $\mathbf{x} \in [0,1]^{24}$ corresponding to the official
learning outcomes. (Table~\ref{tab:perskill}).
Skills are grouped into four progressive groups:
Group~A (S01--S08, class basics),
Group~B (S01--S04, S06--S07, S09--S13, class variables and composition),
Group~C (S01--S04, S06, S09, S14--S21, inheritance and polymorphism), and
Group~D (S01, S14--S15, S22--S24, exception handling). For more details, see Appendix~\ref{sec:appendix-skills}.
Profiles were sampled from 10 archetypes (e.g., ``Absolute Beginner,''
``Lab~2 Proficient,'' ``Advanced'') with Gaussian noise
($\sigma = 0.04$) to produce realistic within-archetype variation.

\paragraph{Assessment slots.}
Every student attempted all 6 assignment slots regardless of skill level, bypassing the adaptive routing to ensure full coverage: Each slot tests a designated subset of the 24 skills; non-applicable skills are marked $-1.0$ in the output vector and excluded from scoring. Scenario entities were assigned deterministically per student (seeded on student ID) to ensure reproducibility.

\paragraph{Generate--then--score protocol.}
For each (student, slot) pair, we executed two sequential API calls to Claude Sonnet 4.6:

\begin{enumerate}
    \item \textbf{Generate}: Given the student's full skill profile
    (24 skill scores with per-skill natural-language descriptors) and
    the assignment question, the model was prompted to produce Python
    code that \emph{precisely matches} the specified skill levels, including
    deliberate errors, omissions, and partial implementations for
    low-scoring skills.
    \item \textbf{Score}: The generated code was submitted to the same
    model's scoring function with the identical rubric, which returned a 24-element observed skill
    vector $\hat{\mathbf{x}} \in \{-1.0\} \cup [0,1]^{24}$ and a scalar
    score $s = \mathrm{round}(\mathrm{mean}(\hat{x}_i : \hat{x}_i \neq -1) \times 100)$.
\end{enumerate}

This yields paired observations $(x_i, \hat{x}_i)$ for every skill $i$
that is applicable in a given slot, providing the raw material for measuring
Equation~\ref{eq:gea}.

\paragraph{Scale.}
All 150 students had completed all 6 slots producing 862 result records and 7{,}788 paired
(true, observed) skill-level observations across 23 of 24 skills
(S13, Dictionary Collection Management, was not tested in any slot). Table~\ref{tab:overall} summarises the aggregate agreement statistics.

\section{GEA Findings}
\label{sec:overall-gea}

\begin{table}[t]
	\centering
	\small
	\caption{Overall GEA statistics (7{,}788 paired skill observations, 150 students, 23 skills). 95\% bootstrap CIs from 1{,}000 resamples.}
	\label{tab:overall}
	\begin{tabular}{lrl}
		\hline
		Metric & Value & 95\% CI \\
		\hline
		Pearson $r$ (pooled) & $0.698$ & $[.684, .712]$ \\
		Mean bias (obs $-$ true) & $+0.059$ & $[+.053, +.066]$ \\
		Exact proficiency match & 34.8\% & \\
		\hline
	\end{tabular}
\end{table}

The pooled Pearson correlation of $r = 0.698$ indicates that the LLM's
evaluative function recovers roughly half the variance
($R^2 \approx 0.49$) in the true skill levels it was asked to generate. The positive mean bias of $+0.059$ confirms the direction predicted by self-preference bias: the model systematically overestimates the skill level of its own generated code.

At the proficiency-level granularity used for reporting (8 ordinal
levels from \emph{Not Demonstrated} to \emph{Mastered}; boundaries in Appendix~\ref{sec:appendix-scale}), exact classification accuracy is only 34.8\%, rising to
64.4\% within $\pm 1$ adjacent level. This represents moderate
agreement, sufficient to distinguish broad skill bands but
insufficient for fine-grained proficiency reporting.

\subsection{Per-Skill GEA}
\label{sec:perskill}

GEA varies dramatically across skills. Table~\ref{tab:perskill} presents the full
per-skill breakdown, sorted by Pearson $r$. After Benjamini-Hochberg correction for multiple comparisons across 23 skills, 19 correlations remain significant at $\alpha = 0.05$; the four non-significant skills are precisely the near-zero GEA tier. Three tiers emerge:

\begin{table}[t]
\centering
\small
\caption{Per-skill GEA metrics, sorted by Pearson $r$.
Skills marked with $\star$ are mandatory in the assessment rubric.}
\label{tab:perskill}
\begin{tabular}{llrrr}
\hline
Skill & & $n$ & $r$ & Bias \\
\hline
S05 & Setter w/ Validation & 290 & .88 & $+$.10 \\
S09$\star$ & Class Variable & 146 & .83 & $+$.26 \\
S23 & Raise Exception & 145 & .80 & $+$.12 \\
S11 & Composition & 139 & .80 & $+$.13 \\
S12 & Delegation & 139 & .80 & $+$.17 \\
S10 & Class Var Mod & 146 & .78 & $+$.14 \\
S18$\star$ & Override Refine & 285 & .77 & $+$.13 \\
S24 & Try/Except & 145 & .77 & $+$.11 \\
S07 & Compute Method & 429 & .72 & $+$.04 \\
S15 & super().\_\_init\_\_ & 433 & .71 & $+$.08 \\
\hline
S06 & \_\_str\_\_ & 717 & .65 & $-$.04 \\
S01 & Class Def & 848 & .62 & $+$.06 \\
S04 & Getter Property & 716 & .59 & $+$.02 \\
S16 & Subclass Attrs & 288 & .57 & $+$.21 \\
S08 & Mutate Method & 287 & .55 & $-$.06 \\
S22 & Custom Exception & 145 & .55 & $+$.21 \\
S02 & Constructor & 717 & .51 & $-$.04 \\
S14 & Subclass Def & 433 & .47 & $+$.25 \\
S03$\star$ & Private Vars & 717 & .43 & $\pm$.00 \\
\hline
S19 & ABC Definition & 114 & .08 & $-$.10 \\
S17$\star$ & Override Replace & 288 & .06 & $+$.19 \\
S21 & Polymorphism & 113 & $-$.03 & $-$.07 \\
S20 & Concrete Subclass & 108 & n/a & $-$.09 \\
\hline
\end{tabular}
\end{table}

\paragraph{Strong GEA ($r > 0.7$; 10 skills).}
Skills with concrete, syntactically verifiable indicators, such as setter
validation logic (S05, $r = 0.88$), class variable declaration with
underscore convention (S09, $r = 0.83$), exception raising (S23,
$r = 0.80$), and composition via object attributes (S11, $r = 0.80$), show
the strongest agreement. These skills have unambiguous code signatures:
a \texttt{@attr.setter} with a conditional check, a \texttt{\_count}
class variable, a \texttt{raise CustomError()} statement. The rubric
criteria map directly to syntactic patterns that both the generator and
evaluator can reliably target.

\paragraph{Moderate GEA ($0.4 < r < 0.7$; 9 skills).}
Foundational skills tested across many slots (S01, S02, S04, S06) show
moderate agreement ($r = 0.43$--$0.65$). These skills are
near-universal in student code (almost every submission defines a
class, writes a constructor, uses properties), creating a ceiling effect
that compresses variance. Subclass-related skills (S14, S16) show
moderate $r$ but high bias ($+0.21$ to $+0.25$), indicating the
generator systematically over-performs on inheritance tasks relative to
the intended skill level.

\paragraph{Near-zero GEA ($r < 0.1$; 4 skills).}
Abstract design skills, including ABC definition (S19, $r = 0.08$), method
overriding by replacement (S17, $r = 0.06$), polymorphism (S21,
$r = -0.03$), and concrete subclass implementation (S20, constant
output), show essentially no correlation between intended and observed
skill levels. S20 is degenerate: the evaluator assigned nearly
identical scores regardless of the true skill level, collapsing all
variation. These skills require \emph{design decisions} (choosing to
define an ABC, choosing to use polymorphic dispatch) rather than
\emph{syntactic patterns}, making them harder for the generator to
``partially implement'' and harder for the evaluator to grade on a
continuous scale.

This pattern (strong GEA for syntactically verifiable skills, weak GEA
for design-level skills) is consistent with the scope limitation noted
in Section~\ref{sec:discussions}: code assessment benefits from partially verifiable ground
truth, but that benefit is concentrated in the syntactic stratum of the
skill taxonomy. Design-level skills behave more like the subjective
assessment domains where GEA is expected to be weakest.

\subsection{Calibration and Bias Structure}
\label{sec:calibration}

\begin{figure}[t]
\centering
\includegraphics[width=\columnwidth]{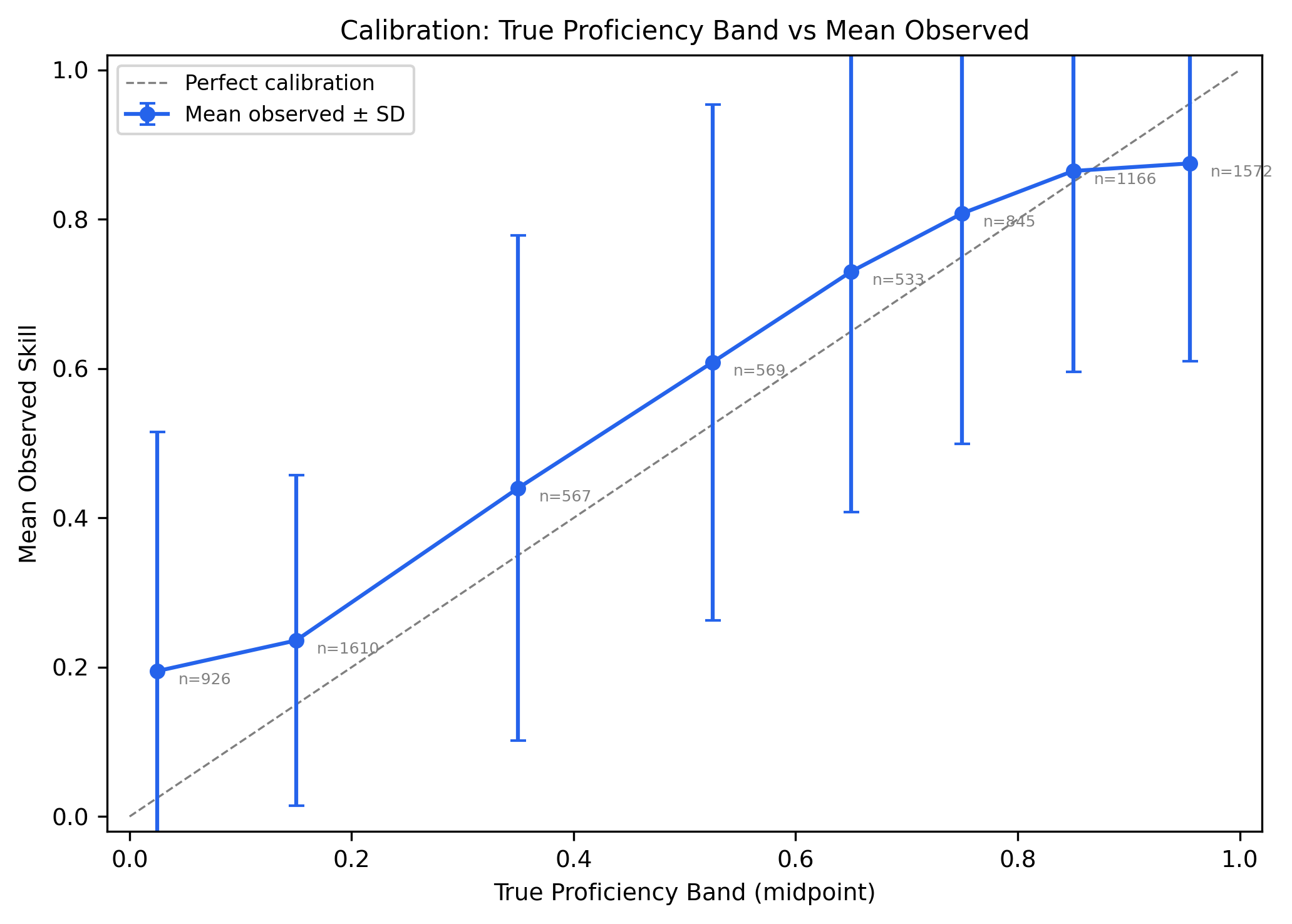}
\caption{Calibration curve: mean observed skill as a function of
true proficiency band. Error bars show $\pm 1$ SD. The dashed
diagonal represents perfect calibration. The curve lies above the
diagonal at low skill levels (overestimation) and converges at high
levels.}
\label{fig:calibration}
\end{figure}

The calibration curve (Figure~\ref{fig:calibration}) reveals an
asymmetric bias:

\begin{itemize}
    \item \textbf{Low-skill overestimation}: Students in the ``Not
    Demonstrated'' band (true $\approx 0.0$) receive mean observed
    scores of $\approx 0.20$, a $+0.20$ bias. The LLM struggles to
    generate authentically incompetent code; even when instructed to
    omit a skill, the generated code tends to include partial or
    vestigial implementations that the evaluator detects.
    \item \textbf{High-skill convergence}: Students in the ``Advanced''
    and ``Mastered'' bands (true $> 0.80$) are scored close to their
    true levels, with the curve approaching the diagonal. The model
    finds it easier to generate competent code and to recognise
    competence.
    \item \textbf{Mid-range compression}: The ``Developing'' through
    ``Proficient'' bands ($0.45$--$0.80$) show the highest variance
    (largest error bars), suggesting the model has difficulty
    maintaining fine-grained distinctions in the middle of the skill
    range.
\end{itemize}

This asymmetry has direct implications for adaptive routing. The
system's routing threshold ($\geq 50$ for High path) operates in the
mid-range where calibration is poorest. The upward bias at low skill
levels means that weak students are systematically overscored, making
them more likely to exceed the threshold and be routed to the High
path, exactly the misrouting scenario described in Section~\ref{sec:discussions}.

\subsection{Proficiency-Level Classification}
\label{sec:confusion}

\begin{figure}[t]
\centering
\includegraphics[width=\columnwidth]{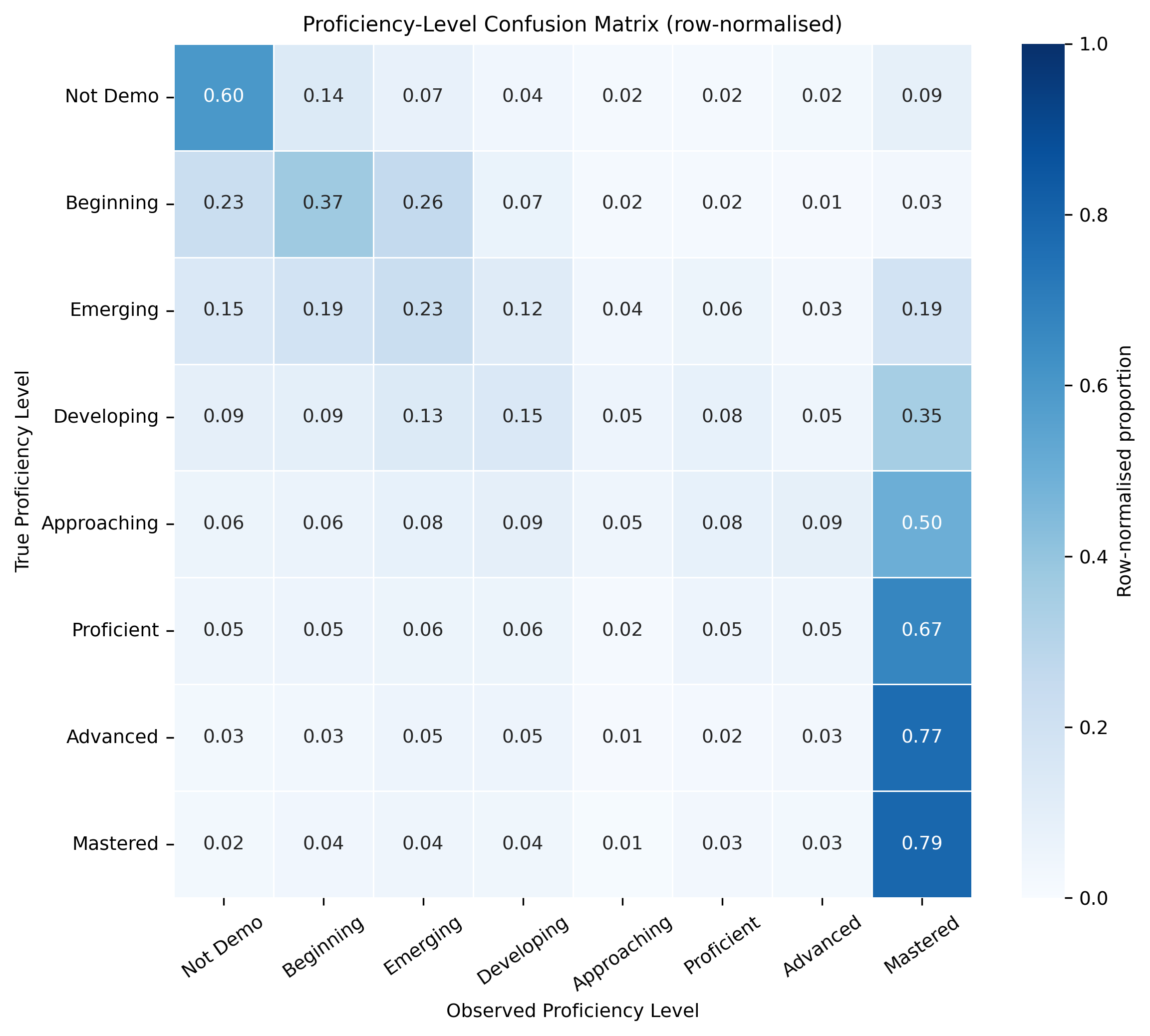}
\caption{Row-normalised confusion matrix mapping true proficiency
levels (rows) to observed proficiency levels (columns). The strong
rightward bias in the ``Approaching'' through ``Mastered'' rows
indicates systematic upward misclassification at higher skill levels.}
\label{fig:confusion}
\end{figure}

The confusion matrix (Figure~\ref{fig:confusion}) reveals the
proficiency-level consequences of the biases identified above.

\paragraph{Upward collapse to ``Mastered.''}
The dominant pattern is a strong rightward shift: students at
``Approaching'' (true $0.60$--$0.70$) are classified as ``Mastered''
50\% of the time; ``Proficient'' (true $0.70$--$0.80$) collapses to
``Mastered'' 67\% of the time; ``Advanced'' (true $0.80$--$0.90$)
reaches 77\%. The evaluator effectively treats any competent
implementation as ``Mastered,'' failing to distinguish gradations within
the upper half of the skill range. This is a direct manifestation of
\textbf{self-preference bias}: the model's own generated
code, even when deliberately degraded, retains low perplexity from the evaluator's perspective, inflating scores toward the ceiling.

\paragraph{Low-skill recovery.}
The ``Not Demonstrated'' level is the best-recovered category at 60\%
exact match, consistent with the observation that the absence of a
skill (e.g., no \texttt{@property} defined at all) is a binary,
syntactically verifiable condition. However, even here, 14\% of truly
absent skills are scored as ``Beginning'' and 9\% as ``Mastered,'' the
latter likely reflecting cases where the generator failed to suppress
the skill despite being instructed to (competence paradox leakage).

\paragraph{Implications for proficiency reporting.}
The 8-level proficiency scale is not reliably recoverable through the
generate-then-score pipeline. A coarser 3- or 4-level scale
would better match the system's effective resolution. For adaptive routing (threshold at 50), the upward bias at low skill levels ($+0.059 \times 100 \approx 6$ points) inflates Stage~1 scores for weak students, increasing misrouting probability to the High path.

\section{Strengthening GEA}
\label{sec:rubrics}

The empirical findings above reveal that GEA is partial and skill-dependent. This section examines the mechanisms available to strengthen it, beginning with the most impactful (granular rubrics) and progressing to complementary strategies that address failure modes rubrics alone cannot resolve.

\subsection{Rubrics as Constrained Reasoning Paths}
\label{sec:rubric-mechanism}

Section~\ref{sec:closed-loop} identified the core architectural problem: generation and evaluation traverse different computational paths through the model with no guarantee of alignment. Granular rubrics address this by providing a \textbf{shared external specification} that constrains both paths to pass through the same intermediate representation, namely the rubric's criteria.

\citet{Wang2025autoscore} formalise this as a graph reasoning problem. Without rubrics, the path from task description to score can traverse arbitrary reasoning states. With rubrics, both generation (``produce code demonstrating these specific skills'') and evaluation (``check for these specific skills'') are forced through the same intermediate checkpoints. Under the assumption that aligning these intermediate nodes increases the overlap between the model's generative and evaluative reasoning paths, rubric-guided assessment should exhibit higher GEA than holistic assessment.

In the context of this system, the rubric defines per-assignment skill vector tables specifying exactly which of the 24 skills are applicable to each assignment and what constitutes mastery. This provides structural decomposition: both the question generator and the response evaluator reference the same granular criteria, reducing the degrees of freedom available for the two paths to diverge. The three-tier GEA pattern in Table~\ref{tab:perskill} provides indirect evidence for this mechanism: skills with unambiguous syntactic rubric criteria (e.g., ``code contains a \texttt{@property} decorator'') show strong GEA, while skills with holistic criteria (e.g., ``demonstrates appropriate use of polymorphism'') show near-zero GEA.

\subsection{Empirical Support from the Literature}
\label{sec:empirical-support}

\textbf{Rubric quality directly predicts scoring accuracy.} \citet{Wu2024scoring} found a Spearman rank correlation of $\rho = 0.94$ ($p < 0.01$) between analytic rubric alignment (with human-crafted rubrics) and automated scoring accuracy. Without rubrics, accuracy was 34.8\%; with human-crafted analytic rubrics, 50.4\%; with LLM-generated rubrics guided by holistic rubrics, 54.6\%. The near-perfect correlation between rubric quality and scoring performance establishes that the evaluative path is strongly anchored by rubric specification. For GEA, this implies that the evaluative side of the agreement can be substantially improved through rubric design alone, without changing the model.

\textbf{Rubric-aligned component extraction yields consistent gains.} AutoSCORE's two-agent design (first extract rubric-relevant components into a structured JSON representation, then score based on those components) improved QWK by up to 37\% (Essay Set, GPT-4o) and 74\% (Science, LLaMA-8B) over single-agent baselines \citep{Wang2025autoscore}. Gains were largest on complex, multi-dimensional rubrics, which parallels this system's 24-skill vector with per-assignment coverage tables. The framework demonstrates that decomposing evaluation into criterion-level checks before holistic scoring reduces rubric misalignment and evaluator shortcuts. This system's architecture already follows this pattern: the evaluator produces a 24-element skill vector before computing a scalar score.

\textbf{Rubrics balance accuracy across proficiency levels.} \citet{Lee2023} found that Chain-of-Thought prompting combined with scoring rubrics yielded a 13.4\% accuracy increase and, critically, more balanced accuracy across different proficiency categories. Without rubrics, LLMs were biased toward certain score ranges; with rubrics, scores distributed more evenly across levels. This directly addresses the calibration asymmetry observed in Figure~\ref{fig:calibration}: rubric-guided evaluation may reduce the systematic overestimation at low skill levels that currently inflates routing scores.

\textbf{Rubrics prevent evaluator shortcuts.} \citet{Wu2024scoring} discovered that providing graded student responses (without rubrics) actually \emph{degraded} rubric alignment; the LLM found superficial keyword shortcuts instead of following the intended reasoning chain. Analytic rubrics prevented this by requiring criterion-level assessment, forcing the model through the intended reasoning path rather than surface-level pattern matching. This finding is particularly relevant to GEA: without rubrics, the evaluator may assign high scores based on surface features (e.g., code length, presence of class definitions) rather than the specific skill indicators the generator was instructed to produce or omit.

Rubrics are therefore a \textbf{necessary but not sufficient} condition for GEA. They are the most practical strategy within a single-model system, but complete GEA assurance requires complementary measures to address failure modes that rubric design alone cannot resolve.

\subsection{Design Principles for GEA-Strengthening Rubrics}
\label{sec:rubric-principles}

Based on the findings above, rubrics that maximise GEA should:
\begin{enumerate}
    \item \textbf{Decompose the construct into discrete, independently assessable skills}, each with a binary or ordinal mastery indicator (as in the 24-skill vector).
    \item \textbf{Specify per-assignment applicability}: explicitly mark which skills are assessed and which are not applicable, so neither generation nor evaluation drifts into unintended territory.
    \item \textbf{Define decision boundaries, not just level descriptions}: specify what distinguishes mastery from non-mastery for each skill, not just what each level ``looks like'' holistically.
    \item \textbf{Use structured output formats}: require the evaluator to produce criterion-level judgements (e.g., JSON skill vectors) before aggregating to a holistic score, following AutoSCORE's component-extraction-then-scoring paradigm \citep{Wang2025autoscore}.
    \item \textbf{Avoid excessive verbosity}: structural clarity outperforms exhaustive description; long rubrics can degrade performance in some models \citep{Yoshida2025}.
\end{enumerate}

\subsection{Complementary Mitigations}
\label{sec:mitigations}

While granular rubrics are the primary strategy for strengthening GEA, they cannot address all failure modes. The competence paradox (the generator producing overly competent code despite low-skill instructions) and self-preference bias (the evaluator inflating scores for model-generated text) require complementary interventions.

\paragraph{Cross-model evaluation.} The most direct intervention is using a different model family to score responses than the one that generated them (e.g., generate with Claude, score with GPT-4). This breaks the self-preference loop whereby a model inflates scores for its own low-perplexity outputs \citep{Panickssery2024}. The perplexity-based mechanism identified by \citet{Wataoka2024} implies that cross-model evaluation should be most beneficial for the upper proficiency levels where self-preference bias is strongest (Figure~\ref{fig:confusion}). Where cross-model evaluation is impractical due to cost or API constraints, \textbf{multi-sample scoring} offers a partial substitute: scoring each response multiple times at non-zero temperature and flagging high-variance items for human review surfaces the stochastic inconsistency that single-pass scoring conceals \citep{Gkeka2025}.

\paragraph{Epistemic state specification.} On the generation side, the competence paradox can be mitigated by moving beyond naive role-prompting (``act as a beginner''). \citet{Yuan2026} propose using structured misconception inventories and knowledge component graphs that constrain the generative path to produce behaviourally realistic responses. Rather than asking the model to simulate a general proficiency level, the prompt specifies which knowledge components the student has and has not acquired, which common misconceptions are active, and which error patterns should appear. This converts a vague instruction (``produce beginner code'') into a concrete specification that the generator can follow more reliably, directly improving the generative side of GEA.

\paragraph{Real-student pilot validation.} Simulation-derived thresholds should be validated against a small real-student cohort before operational use, following standard psychometric practice. Even a pilot of 10--20 students on a subset of skills would provide an external anchor against which to calibrate the simulation's bias estimates. This is particularly important for the routing threshold, where the $+6$-point upward bias identified in Section~\ref{sec:calibration} may require adjustment before deployment.

\section{Discussion}
\label{sec:discussions}

\paragraph{Threshold sensitivity.} We swept the Stage~1 routing threshold $\theta$ from 30 to 70 (Table~\ref{tab:threshold}), re-routing 140 students with complete data (10 excluded due to incomplete Stage~2 records).

\begin{table}[h]
	\centering\small
	\begin{tabular}{rrrrrr}
		\hline
		$\theta$ & Flip\% & Adv\% & Int\% & Beg\% & Mis\% \\
		\hline
		30 & 10.7 & 66.4 & 26.4 &  7.1 & 45.0 \\
		40 &  7.1 & 43.6 & 45.0 & 11.4 & 41.4 \\
		\textbf{50} & \textbf{0.0} & \textbf{24.3} & \textbf{60.0} & \textbf{15.7} & \textbf{34.3} \\
		60 &  5.7 & 15.7 & 62.1 & 22.1 & 28.6 \\
		70 & 18.6 &  8.6 & 56.4 & 35.0 & 18.6 \\
		\hline
	\end{tabular}
	\caption{Threshold sensitivity sweep. Flip\%: routing changes vs.\ baseline $\theta{=}50$. Mis\%: fraction misaligned with true archetype ability.}
	\label{tab:threshold}
\end{table}

A \textbf{stability plateau} spans $\theta \in [45, 55]$ ($<$5\% flips). Misclassification decreases monotonically as $\theta$ rises (45\% at $\theta{=}30$ to 19\% at $\theta{=}70$) because the positive scoring bias pushes observed scores above true ability. However, at $\theta{=}70$ nearly one in five students would be reclassified. The baseline $\theta{=}50$ represents a pragmatic compromise within the stability plateau.

\paragraph{Decomposing GEA failure.} When GEA is low, the deficit could stem from the generator, the evaluator, or both. For syntactically verifiable skills (e.g., S05, $r = 0.88$), unambiguous code signatures constrain both paths. For design-level skills (e.g., S21, $r = -0.03$), the generator likely over-produces while the evaluator lacks binary markers to assess degree. Isolating each component requires human scoring of generated code or human-written code at specified skill levels.

\paragraph{Domain dependence.} Code assessment occupies a privileged position because programming tasks have partially verifiable ground truth: a class either defines a \texttt{@property} or it does not. Subjective domains (essay argumentation, creative writing) lack these anchors, so GEA findings here likely represent an \textbf{upper bound}. The rubric-as-config architecture is domain-agnostic, but empirical GEA guarantees are domain-specific.

\paragraph{Model scaling.} We repeated the full simulation using Haiku~4.5 with identical rubrics and profiles (Table~\ref{tab:model-scaling}).

\begin{table}[h]
	\centering\small
	\begin{tabular}{lcc}
		\hline
		\textbf{Metric} & \textbf{Haiku 4.5} & \textbf{Sonnet 4.6} \\
		\hline
		Signed bias (obs$-$true)& $+$0.31 & $+$0.06 \\
		Pearson $r$ (record-level)& 0.64  & 0.92 \\
		Terminal Advanced (\%)  & 88.0  & 24.1 \\
		\hline
	\end{tabular}
	\caption{Model scaling comparison. Pearson $r$ is computed at the \emph{record level} ($n = 862$ assignment records), which averages across ${\sim}9$ skills per record and is therefore higher than the pooled skill-level $r = 0.698$ in Table~\ref{tab:overall}.}
	\label{tab:model-scaling}
\end{table}

Haiku inflates scores by $+17.6$ points on average, scoring Absolute Beginners at $44.8$ (vs.\ Sonnet's $19.8$), above the routing threshold, and assigning 88\% of students to Advanced. The pooled $r$ difference (Sonnet 0.698 vs.\ Haiku 0.447) is significant (Fisher $z = 23.8$, $p < 10^{-100}$). GEA is therefore \textbf{scale-dependent}: self-preference bias and the competence paradox are amplified at smaller scale.

\section{Conclusion}
\label{sec:conclusion}

We introduced Generative-Evaluative Agreement (GEA) as a necessary validity criterion for LLM-enabled adaptive assessment. Using 150 synthetic profiles on a Python OOP assessment with Claude Sonnet~4.6, the model recovers roughly half the intended skill variance ($r = 0.698$, 95\% CI $[.684, .712]$) with systematic positive bias, and GEA is strongly skill-dependent: high for syntactically verifiable skills, near zero for design-level skills.

\section*{Limitations}
The reported GEA estimates should be read with several scoping constraints in mind. All 150 profiles are LLM-sampled rather than drawn from real students; authentic learner errors likely differ from simulated ones, so a pilot of 10--20 real students would provide an external anchor for the bias estimates and a check on the simulated-error distribution. We evaluate two Claude models (Sonnet~4.6, Haiku~4.5) on Python OOP code only; cross-family replication (e.g., GPT-4o, Gemini, open-weight models) and subjective domains that lack the partial verifiability of code (essays, open-ended argumentation) would test whether the strong-vs-weak-GEA stratification we observe generalises beyond this setting. The rubric-decomposition argument and complementary mitigations (cross-model scoring, epistemic-state specification, multi-sample evaluation) are supported by prior work and by the per-skill GEA contrast in Table~\ref{tab:perskill}, but we do not directly ablate rubric granularity or scorer identity on the same task; comparing holistic against decomposed rubrics, and same-model against cross-model scoring, is the most immediate empirical follow-up. Finally, GEA can in principle fail to detect bias when generator and evaluator share an identical distortion---for example, a sycophantic scorer that deflates an inflated generator's score to match user expectations---and precisely quantifying this residual risk requires human-scored anchor responses to decompose GEA failure between the two functions (Section~\ref{sec:discussions}). The intended skill level $x$ nonetheless remains an external reference that distinguishes GEA measurement from pure closed-loop self-validation, and GEA provides a concrete, measurable criterion any LLM-based assessment system can report before deployment.

\bibliography{custom}

\appendix

	\section{Assessment Architecture}
	\label{sec:appendix-architecture}

	The system is a conversational assessment tool delivered via Telegram Bot that conducts adaptive, scenario-based coding assessments. Students work through multiple progressive mini coding assignments per stage, evaluated by Claude AI against predefined rubrics. The system routes students to a terminal difficulty level based on cumulative stage performance.

	\paragraph{Two-stage adaptive routing.}
	All students begin with Stage~1 (2 assignments covering class basics and composition). Based on their cumulative Stage~1 score relative to threshold~$\theta$, students are routed to either the \emph{High Performer} path (inheritance and exception handling) or the \emph{Low Performer} path (reinforcement of foundational skills). After Stage~2, a terminal level is assigned:

	\begin{itemize}
		\item \textbf{Advanced}: Stage~2 High path, cumulative $\geq \theta$
		\item \textbf{Intermediate}: Stage~2 High path, cumulative $< \theta$; \emph{or} Stage~2 Low path, cumulative $\geq \theta$
		\item \textbf{Beginner}: Stage~2 Low path, cumulative $< \theta$
	\end{itemize}

	\paragraph{Dynamic question generation.}
	Questions are not pre-stored. Claude generates each assignment dynamically at runtime based on the rubric, the current stage/path, and a scenario template. Each student receives a slightly different variation of the same assignment---same core objective and difficulty, but different scenario entities (e.g., bank account, airline, cinema), discouraging copying. Every assignment begins with an ASCII UML class diagram showing the class(es) the student must implement.

	\paragraph{Scoring.}
	Each submission is evaluated by Claude against the rubric for that assignment. The model returns a 24-element skill vector $\hat{\mathbf{x}} \in \{-1.0\} \cup [0,1]^{24}$ (where $-1.0$ denotes inapplicable skills) and a scalar score $s = \mathrm{round}(\mathrm{mean}(\hat{x}_i : \hat{x}_i \neq -1) \times 100)$. The scalar score drives routing; the skill vector provides diagnostic detail.

	\section{Skill Taxonomy}
	\label{sec:appendix-skills}

	The 24 skills are organised into four progressive groups corresponding to the course's lab sequence. Skills marked with $\star$ are mandatory anchors in the assessment rubric.

	\begin{table*}[h]
		\centering\small
		\caption{Complete 24-skill taxonomy with descriptions and demonstration criteria.}
		\label{tab:skill-taxonomy}
		\begin{tabular}{lp{3.2cm}p{5.5cm}p{4.5cm}}
			\hline
			ID & Skill & Description & Demonstrated by \\
			\hline
			\multicolumn{4}{l}{\textbf{Group A --- Class Basics (Lab 1)}} \\
			S01 & Class Definition & Defines a class using \texttt{class} with PascalCase name; non-empty body & Any valid class definition with $\geq$1 method \\
			S02 & Constructor Init & Uses \texttt{\_\_init\_\_(self, ...)} to initialise instance attributes & \texttt{\_\_init\_\_} present, $\geq$2 attrs assigned via \texttt{self} \\
			S03$\star$ & Private Vars & Declares \texttt{self.\_\_attr} with name mangling & $\geq$1 \texttt{self.\_\_attr} with property/getter access \\
			S04 & Getter Property & \texttt{@property} decorator exposes private attribute & Method with \texttt{@property} returning \texttt{self.\_\_attr} \\
			S05 & Setter w/ Validation & \texttt{@attr.setter} with $\geq$1 validation rule & Setter checks condition before assigning \\
			S06 & \texttt{\_\_str\_\_} & Returns human-readable string using instance data & f-string with $\geq$2 instance attributes \\
			S07 & Compute Method & Instance method reading \texttt{self} attrs, returns computed value & \texttt{return} using \texttt{self.attr} in calculation \\
			S08 & Mutate Method & Instance method modifying \texttt{self} attributes & Assigns new value to \texttt{self.\_\_attr} \\
			\hline
			\multicolumn{4}{l}{\textbf{Group B --- Class Variables \& Composition (Lab 2)}} \\
			S09$\star$ & Class Variable & Shared class-level \texttt{\_attr = value} (not in \texttt{\_\_init\_\_}) & \texttt{\_attr} at class level, accessed via \texttt{ClassName.\_attr} \\
			S10 & Class Var Mod & Correctly accesses/modifies class variable at runtime & Change reflected across all instances \\
			S11 & Composition & Has-a relationship: object stored as attribute (DI) & \texttt{\_\_init\_\_} accepts object param, assigns to \texttt{self.\_\_other} \\
			S12 & Delegation & Outer class calls composed object's methods/properties & \texttt{self.\_\_inner.method()} in outer class \\
			S13 & Dict Collection & Dictionary manages collection with add/search/remove & \texttt{dict[key] = obj}, \texttt{dict.get()}, \texttt{dict.pop()} \\
			\hline
			\multicolumn{4}{l}{\textbf{Group C --- Inheritance (Lab 3)}} \\
			S14 & Subclass Def & \texttt{class Child(Parent):} with correct parent & Subclass instance can call parent methods \\
			S15 & \texttt{super().\_\_init\_\_()} & Calls \texttt{super().\_\_init\_\_(...)} first in subclass & First statement in child \texttt{\_\_init\_\_} \\
			S16 & Subclass Attrs & New attributes added after \texttt{super()} call & $\geq$1 \texttt{self.\_\_new\_attr} after \texttt{super()} \\
			S17$\star$ & Override Replace & Completely replaces parent method (no \texttt{super()}) & Same-name method with entirely new logic \\
			S18$\star$ & Override Refine & Calls \texttt{super().method()} then extends result & \texttt{super().method()} + additional logic \\
			S19 & ABC Definition & \texttt{ABC} + \texttt{@abstractmethod} to define contract & \texttt{from abc import ABC, abstractmethod} \\
			S20 & Concrete Subclass & Implements all abstract methods with meaningful bodies & Subclass instantiates without \texttt{TypeError} \\
			S21 & Polymorphism & Same method called on mixed-type list, no type checks & Loop over heterogeneous list, no \texttt{isinstance} \\
			\hline
			\multicolumn{4}{l}{\textbf{Group D --- Exception Handling (Lab 4)}} \\
			S22 & Custom Exception & Class inheriting from \texttt{Exception} & \texttt{class MyException(Exception): pass} \\
			S23 & Raise Exception & \texttt{raise} custom exception with descriptive message & \texttt{if condition: raise MyException("msg")} \\
			S24 & Try/Except & \texttt{try/except} catching custom + built-in exceptions & $\geq$2 distinct conditions caught with messages \\
			\hline
		\end{tabular}
	\end{table*}

	\section{Assessment Slot--Skill Coverage}
	\label{sec:appendix-slots}

	Table~\ref{tab:slot-skills} shows which skills are assessed (scored $0.0$--$1.0$) in each of the 6 assessment slots. Skills not listed for a slot receive $-1.0$ (not applicable) in the skill vector.

	\begin{table*}[h]
		\centering\small
		\caption{Slot-to-skill mapping. Each slot tests a designated subset of the 24 skills; remaining skills are marked $-1.0$.}
		\label{tab:slot-skills}
		\begin{tabular}{llll}
			\hline
			Slot & Content & Skills assessed & $n$ skills \\
			\hline
			Stage~1, A1 & Lab~1: classes \& properties & S01--S08 & 8 \\
			Stage~1, A2 & Lab~2: composition & S01--S04, S06--S07, S11--S12 & 8 \\
			Stage~2 High, A1 & Lab~3: inheritance & S01--S04, S06, S09--S10, S14--S21 & 15 \\
			Stage~2 High, A2 & Lab~4: exceptions & S01, S14--S15, S22--S24 & 6 \\
			Stage~2 Low, A1 & Lab~1 reinforcement & S01--S08 & 8 \\
			Stage~2 Low, A2 & Lab~3 introduction & S01--S04, S06, S14--S18 & 10 \\
			\hline
		\end{tabular}
	\end{table*}

	\section{Proficiency Scale}
	\label{sec:appendix-scale}

	Continuous skill scores are mapped to ordinal proficiency levels using the boundaries in Table~\ref{tab:proficiency-scale}. These levels are used for the confusion matrix analysis (Section~\ref{sec:confusion}) and for natural-language descriptors in student profiles.

	\begin{table}[h]
		\centering\small
		\caption{Proficiency level boundaries.}
		\label{tab:proficiency-scale}
		\begin{tabular}{llc}
			\hline
			Level & Score range & Midpoint \\
			\hline
			Not Demonstrated & $[0.00, 0.05)$ & 0.025 \\
			Beginning & $[0.05, 0.25)$ & 0.15 \\
			Emerging & $[0.25, 0.45)$ & 0.35 \\
			Developing & $[0.45, 0.60)$ & 0.525 \\
			Approaching & $[0.60, 0.70)$ & 0.65 \\
			Proficient & $[0.70, 0.80)$ & 0.75 \\
			Advanced & $[0.80, 0.90)$ & 0.85 \\
			Mastered & $[0.90, 1.00]$ & 0.95 \\
			\hline
		\end{tabular}
	\end{table}

	\section{Student Profile Archetypes}
	\label{sec:appendix-archetypes}

	150 synthetic student profiles were sampled from 10 archetypes (Table~\ref{tab:archetypes}). Each archetype defines per-group skill ranges from which individual skill scores are drawn uniformly, with Gaussian noise ($\sigma = 0.04$) added to produce within-archetype variation. Profiles were generated with a fixed random seed for reproducibility.

	\begin{table*}[h]
		\centering\small
		\caption{Archetype definitions. Each cell shows the $[\text{lo}, \text{hi}]$ range for uniform sampling of skill scores within that sub-group. Sub-groups: A = S01--S08 (class basics), B = S09--S13 (composition), C\textsubscript{1} = S14--S16 (basic subclass), C\textsubscript{2} = S17--S18 (mandatory overrides), C\textsubscript{3} = S19--S21 (advanced inheritance), D = S22--S24 (exceptions).}
		\label{tab:archetypes}
		\begin{tabular}{lrcccccc}
			\hline
			Archetype & \% & A & B & C\textsubscript{1} & C\textsubscript{2} & C\textsubscript{3} & D \\
			\hline
			Absolute Beginner   &  8 & .00--.22 & .00--.10 & .00--.07 & .00--.07 & .00--.05 & .00--.05 \\
			Lab 1 Developing    & 12 & .22--.52 & .00--.14 & .00--.09 & .00--.09 & .00--.05 & .00--.05 \\
			Lab 1 Proficient    & 15 & .52--.78 & .00--.22 & .00--.14 & .00--.12 & .00--.07 & .00--.05 \\
			Lab 1 Mastered      & 12 & .78--1.0 & .18--.48 & .00--.18 & .00--.14 & .00--.09 & .00--.05 \\
			Lab 2 Developing    & 12 & .68--1.0 & .32--.62 & .00--.18 & .00--.14 & .00--.09 & .00--.07 \\
			Lab 2 Proficient    & 10 & .78--1.0 & .62--.92 & .08--.28 & .05--.22 & .00--.12 & .00--.09 \\
			Lab 3 Developing    & 12 & .72--1.0 & .62--1.0 & .28--.62 & .22--.58 & .00--.22 & .00--.14 \\
			Lab 3 Proficient    & 10 & .82--1.0 & .72--1.0 & .58--.88 & .52--.85 & .08--.38 & .00--.18 \\
			Lab 4 Developing    &  5 & .78--1.0 & .72--1.0 & .68--1.0 & .62--.92 & .18--.58 & .28--.62 \\
			Advanced            &  4 & .85--1.0 & .80--1.0 & .78--1.0 & .75--1.0 & .52--.92 & .68--1.0 \\
			\hline
		\end{tabular}
	\end{table*}

	\paragraph{Example profile.}
	Table~\ref{tab:example-profile} shows a representative ``Lab~2 Developing'' profile. The student has strong class basics (Group~A, mean $0.84$), partial composition skills (Group~B, mean $0.46$), and minimal inheritance/exception knowledge (Groups~C--D, $< 0.10$). Each skill carries a natural-language descriptor (shown for selected skills) that is included in the generation prompt to constrain the LLM's code output.

	\begin{table}[h]
		\centering\small
		\caption{Excerpt from student profile 0114 (Lab~2 Developing archetype, overall score 0.41).}
		\label{tab:example-profile}
		\begin{tabular}{llrl}
			\hline
			Skill & Name & Score & Level \\
			\hline
			S01 & Class Definition & 0.82 & Advanced \\
			S03$\star$ & Private Vars & 0.70 & Proficient \\
			S05 & Setter w/ Validation & 0.92 & Mastered \\
			S09$\star$ & Class Variable & 0.37 & Emerging \\
			S11 & Composition & 0.66 & Approaching \\
			S14 & Subclass Def & 0.15 & Beginning \\
			S17$\star$ & Override Replace & 0.00 & Not Dem. \\
			S22 & Custom Exception & 0.01 & Not Dem. \\
			\hline
		\end{tabular}
	\end{table}

	\section{Prompt Templates}
	\label{sec:appendix-prompts}

	\subsection{Code Generation Prompt (Simulation)}
	\label{sec:appendix-codegen}

	For each (student, slot) pair, the following prompt is sent to Claude Sonnet~4.6 to generate code that matches the student's skill profile. The \texttt{STUDENT SKILL PROFILE} block lists only the skills applicable to the current slot, with each skill's numeric score, proficiency level, and natural-language descriptor.

	\begin{quote}\small\ttfamily
	You are simulating a student submitting a Python OOP coding assignment.\\[4pt]
	The student's exact skill profile for the skills tested in this assignment is provided below.
	Write Python code that a student at PRECISELY these skill levels would produce.
	Faithfully reflect each described weakness and strength --- do not average them out or homogenise the code.\\[4pt]
	STUDENT SKILL PROFILE (relevant skills only):\\
	~~- S01 Class Definition: 0.82 (Advanced) --- Class is well-defined, correctly named, and structurally complete with multiple methods.\\
	~~- S03 Private Instance Variables: 0.70 (Proficient) --- self.\_\_attr used for key attributes with property getters; not all attributes are private.\\
	~~- ...\\[4pt]
	ASSIGNMENT:\\
	\{the generated assignment text\}\\[4pt]
	Rules:\\
	1. Output Python code only --- no explanations, no markdown fences, no preamble.\\
	2. For skills rated ``Not Demonstrated'' or ``Beginning'', the code must clearly exhibit the described gap.\\
	3. For skills rated ``Advanced'' or ``Mastered'', that aspect of the code must be correct and complete.\\
	4. Each skill reflects its own level independently --- the code can be strong in one area and weak in another.\\
	5. Use realistic student-style naming and formatting consistent with the described skill levels.
	\end{quote}

	\subsection{Scoring Prompt (Evaluation)}
	\label{sec:appendix-scoring}

	The scoring function sends the full rubric document (including per-assignment skill vector tables with scoring guidance) along with the student's submission. The prompt instructs the model to:

	\begin{quote}\small\ttfamily
	You are a coding assessment scorer for a Python OOP course.\\[4pt]
	The full rubric is below.\\[4pt]
	RUBRIC:\\
	\{full RUBRICS.md content\}\\[4pt]
	---\\[4pt]
	Score the student's submission for:\\
	- Stage: \{stage\}~~- Path: \{path\}~~- Assignment: \{n\} of 2\\[4pt]
	Assignment given to the student:\\
	"""\{question text\}"""\\[4pt]
	Student's submission:\\
	"""\{student code\}"""\\[4pt]
	Instructions:\\
	1. Locate the rubric section for this stage, path, and assignment number.\\
	2. Fill in the 24-element skill\_vector exactly as defined in the Skill Vector table.\\
	~~~- Use -1.0 for skills marked -1.0 (not applicable).\\
	~~~- Use a float 0.0--1.0 for all other skills, following the scoring guidance.\\
	~~~- Use intermediate values (e.g.\ 0.3, 0.7) freely.\\
	3. Compute score = round(mean(v\_i for v\_i if v\_i != -1.0) * 100).\\
	4. Write 2--4 sentences of constructive feedback.\\[4pt]
	Return ONLY a valid JSON object:\\
	\{``score'': <int>, ``feedback'': ``<text>'', ``skill\_vector'': [<s01>, ..., <s24>]\}
	\end{quote}

	\subsection{Question Generation Prompt}
	\label{sec:appendix-qgen}

	The question generator receives the full rubric and is instructed to locate the correct section for the given stage, path, and assignment number, substitute the student's scenario entity into the template, produce the class diagram followed by coding instructions, and return only the assignment text the student will see. Scenario entities are drawn from pre-defined lists:

	\begin{itemize}
		\item \textbf{Stage~1}: bank account, airline booking, cinema, grade tracker, event planner, inventory, point-of-sale, membership
		\item \textbf{Stage~2 High}: banking hierarchy, airline, vehicle fleet, cinema chain, course catalogue, training programme
		\item \textbf{Stage~2 Low}: contact book, recipe manager, budget tracker, shopping list, event log, book collection
	\end{itemize}

	Entities are assigned deterministically per student (seeded on student ID) to ensure reproducibility across simulation restarts.

	\section{Rubric Excerpts: Strong vs.\ Weak GEA Skills}
	\label{sec:appendix-rubric-contrast}

	To illustrate the rubric specificity hypothesis (Section~\ref{sec:rubric-mechanism}), we present the scoring guidance for a strong-GEA skill and a near-zero-GEA skill as they appear in the rubric's skill vector tables.

	\paragraph{S05 --- Setter with Validation ($r = 0.88$, strong GEA).}
	The rubric specifies concrete syntactic markers at each score level:

	\begin{quote}\small
	\textbf{Scoring guidance}: 1.0 = \texttt{@attr.setter} with a meaningful validation rule; 0.5 = setter present but no validation logic; 0.0 = absent.
	\end{quote}

	\noindent The 8-level proficiency descriptors used in student profiles further constrain generation:

	\begin{table}[h]
		\centering\small
		\caption{S05 proficiency descriptors (excerpt).}
		\begin{tabular}{lp{5.5cm}}
			\hline
			Level & Descriptor \\
			\hline
			Not Dem. & No setter defined; private attributes cannot be updated after construction. \\
			Emerging & Setter present but validation logic is absent; any value is accepted. \\
			Proficient & Setter enforces a clear validation rule and handles the invalid case appropriately; minor gap. \\
			Mastered & Setter enforces thorough validation with appropriate response; all edge cases covered. \\
			\hline
		\end{tabular}
	\end{table}

	The binary nature of the criterion (setter with validation present or absent) provides an unambiguous code signature that both the generator and evaluator can reliably target.

	\paragraph{S21 --- Polymorphism via Shared Interface ($r = -0.03$, near-zero GEA).}
	The rubric criterion is holistic rather than syntactic:

	\begin{quote}\small
	\textbf{Scoring guidance}: 1.0 = loop/function calls same method on mixed-type list without \texttt{isinstance} checks; 0.5 = loop present but uses type checks; 0.0 = no polymorphic usage.
	\end{quote}

	\noindent The corresponding proficiency descriptors:

	\begin{table}[h]
		\centering\small
		\caption{S21 proficiency descriptors (excerpt).}
		\begin{tabular}{lp{5.5cm}}
			\hline
			Level & Descriptor \\
			\hline
			Not Dem. & No polymorphic usage; objects are type-checked before any method call. \\
			Emerging & A polymorphic call is made but only one subclass type is present. \\
			Proficient & Multiple subclass types in a collection; same method called without type checks; each responds correctly. \\
			Mastered & Polymorphism fully and cleanly demonstrated; shared interface called on mixed-type collection with no type checks and correct output per type. \\
			\hline
		\end{tabular}
	\end{table}

	Polymorphism is a \emph{design choice} (choosing to iterate over a heterogeneous list) rather than a syntactic pattern, making it harder for the generator to ``partially implement'' and harder for the evaluator to grade on a continuous scale. The descriptor levels describe \emph{degrees of design completeness} rather than presence/absence of identifiable code tokens.

	\section{Routing Logic}
	\label{sec:appendix-routing}

	The routing state machine operates as follows. Let $\bar{s}_k$ denote the cumulative (mean) score after completing all assignments in stage~$k$.

	\begin{enumerate}
		\item \textbf{Stage~1 completion}: Compute $\bar{s}_1$. If $\bar{s}_1 \geq \theta$, route to Stage~2 High path; otherwise Stage~2 Low path.
		\item \textbf{Stage~2 completion}: Compute $\bar{s}_2$.
		\begin{itemize}
			\item High path: if $\bar{s}_2 \geq \theta$ then \emph{Advanced}; else \emph{Intermediate}.
			\item Low path: if $\bar{s}_2 \geq \theta$ then \emph{Intermediate}; else \emph{Beginner}.
		\end{itemize}
	\end{enumerate}

	The threshold $\theta = 50$ is a placeholder pending real-student calibration. The threshold sensitivity analysis (Table~\ref{tab:threshold}) confirms a stability plateau at $\theta \in [45, 55]$ where $<$5\% of routing decisions change.

	Each stage comprises exactly 2 assignments. Scores within a stage are averaged (not summed), so $\bar{s}_k = \frac{1}{2}\sum_{j=1}^{2} s_{k,j}$ where $s_{k,j} \in [0, 100]$. Sessions are stored in memory keyed by Telegram user ID and do not survive bot restarts.

	
\end{document}